# Machine-Learning Identification of Hemodynamics in Coronary Arteries in the Presence of Stenosis

Mohammad Farajtabar[2], M Momeni[3], Mohit Biglarian[5], Morteza Miansari[1,4]*

[1]Micro+Nanosystems & Applied Biophysics Laboratory, Department of Mechanical Engineering, Babol Noshirvani University of Technology, P.O. Box 484, Babol, Iran

[2]Department of Advanced Robotics, Italian Institute of Technology, Genoa, Italy

[3]Department of Mechanical Engineering, Babol Noshirvani University of Technology, Babol Noshirvani University of Technology, P.O. Box 484, Babol, Iran

[4]Department of Cancer Medicine, Cell Science Research Center, Royan Institute for Stem Cell Biology and Technology, ACECR, Isar 11, 47138-18983 Babol, Iran

[5]Department of Mechanical Engineering, Sharif University of Technology, Tehran, Iran

---

* Corresponding Author: Morteza Miansari, mmiansari@nit.ac.ir

**Abstract**

Prediction of the blood flow characteristics is of utmost importance for understanding the behavior of the blood arterial network, especially in the presence of vascular diseases such as stenosis. Computational fluid dynamics (CFD) has provided a powerful and efficient tool to determine these characteristics including the pressure and velocity fields within the network. Despite numerous studies in the field, the extremely high computational cost of CFD has led the researchers to develop new platforms including Machine Learning approaches that instead provide faster analyses at a much lower cost. In this study, we put forth a Deep Neural Network framework to predict flow behavior in a coronary arterial network with different properties in the presence of any abnormality like stenosis. To this end, an artificial neural network (ANN) model is trained using synthetic data so that it can predict the pressure and velocity within the arterial network. The data required to train the neural network were obtained from the CFD analysis of several geometries of arteries with specific features in ABAQUS software. Blood pressure drop caused by stenosis, which is one of the most important factors in the diagnosis of heart diseases, can be predicted using our proposed model knowing the geometrical and flow boundary conditions of any section of the coronary arteries. The efficiency of the model was verified using three real geometries of LAD's vessels. The proposed approach precisely predicts the hemodynamic behavior of the blood flow. The average accuracy of the pressure prediction was 98.7% and the average velocity magnitude accuracy was 93.2%. The key contribution of the proposed framework is that it can predict the hemodynamics of the blood flow in any segment of the arterial network knowing only geometrical features and velocity boundary conditions of the segment which are accessible using non-invasive methods. According to the results of testing the model on three patient-specific geometries, model can be considered as an alternative to finite element methods as well as other hard-to-implement and time-consuming numerical simulations. Furthermore, although, as a cost and time-efficient method, the model was trained for coronary arteries, however, it can be easily retrained for other types of arteries in the body using a new dataset.

**Keywords**: Deep neural networks; Flow modeling; Arterial Network; Stenosis; Hemodynamics

1. Introduction

Biomedical engineering as a new area of engineering follows the understanding of the blood flow and human organs under different circumstances. Progress in clinical tests and numerical simulation techniques has opened new avenues toward monitoring the human cardiovascular system from different aspects such as disease surveys, image processing, modeling and numerical simulations [1-4], and multiphysics [5, 6]. Bioengineer scientists have used and demonstrated computational fluid dynamic (CFD) capability as a powerful tool for the investigation of physiological flows [7-9], diagnosis and treatment as well as designing new devices for clinical tests [14-16]. Among all, there has been a special focus on the modeling of the cardiovascular system such as simulation of blood flow in Coronary arteries [17-20], stenosis and atherosclerosis [17-20], simulation of LDL and monocyte depositions as a critical stage in plaque generation and designing and construction of the treatment tools and methods such as in-vitro stents [21,22].

Although the computational modeling and prototyping using new methods of imaging, image processing, and especially 3-D printing have significantly boosted the development of cardiovascular medicine, however, their implantation has always been limited. This is mainly because of the patient-specified nature of cardiovascular diseases, the design and construction of geometry and simulation for each patient is a very time-consuming and computationally expensive process. Moreover, model parameterization and its requirements to the knowledge of physiological metrics such as microcirculatory resistance in coronary arteries is another challenging issue. It requires numerous simulations and computational modeling for understanding the relative importance of the physiological parameters and determination of the most influential parameters, and those that can be ignored or averaged. All together can lead to data sets that are too large and/or complex to be dealt with by conventional data processing platforms. Furthermore, processing and evaluating the vast volume of data collected is not only a time-consuming and labor-intensive process, but it is often vulnerable to poor accuracy and user bias. Fortunately, advanced computerized techniques and algorithms, such as Machine Learning (ML), as a subdivision of Artificial Intelligence (AI), have made it possible to handle large data without using special assumptions about the simulated system and managing the challenges associated with the computational modeling. More recently, ML has been developed based on a deep learning model, which is capable of estimating the fluid flow in different bioapplications, especially in cancers, diagnosis, and treatment with high accuracy [23-25].

Training a machine based on deep learning can help physicians and medicines via developing a high accuracy AI system that predicts hemodynamic characteristics of the blood in cardiovascular systems [26-27]. Sankaran et al [28] used a machine learning approach to estimate the simulation-based solution for the investigation of the impact of geometric uncertainty on hemodynamic of blood flow. Itu et al [29] performed a study on the use of machine-learning for fractional flow reserve (FFR) prediction in coronary as an alternative to physics-based methods. The trained model predicted FFR for every point at the centerline of the coronary arteries for 87 patients and 125 lesions. The results showed an excellent agreement between machine-learning and physics-based predictions. Wu et al [30] used machine learning and two models of neural networks for the prediction of systolic pressure in coronary arteries. They used near 500 cases that 80% were used for training the machine and 20% were used to test and investigate the effect of lifestyle parameters such as age, exercise, stress level, and sexuality on blood pressure. They presented this model as an early warning system for cardiovascular diseases. Narang et al [31] for the first time performed a study on automated ML-based analysis of 3DE data sets for left heart chamber volumes. They illustrated that their method allows dynamic measurement of LV and LA volumes throughout the cardiac cycle as well as estimation of useful ejection and filling indices with high accuracy.

Tesche et al [32] compared a machine learning-based prediction of FFR in coronary with an approach based on computational fluid dynamics modeling. They studied their respective diagnostic performance against stenosis grading at coronary CT angiography. Coenen et al [33] performed a machine learning study on stenosis in coronary and showed that without CFD-based simulations, onsite CT-FFR based on Machine learning can predict hemodynamic severity of coronary stenosis with very high accuracy. Kissas et al [34] presented a study on the use of one-dimensional models of pulsatile blood flow for the construction of deep neural networks. This constructed machine estimated values that were not possible to be reliably measured in a non-invasive manner (e.g., blood pressure), by use of the governing equations of fluid flow with scattered measurements by medical images, without any need for conventional computational models. Plaques rupture as main parameters in the majority of acute coronary syndromes has been considered and studied in Cilla et al [35]. Ruptures are closely associated with stress concentrations for which they developed a machine to help the clinical professionals on making decisions of the vulnerability of the atheroma plaque and presented two potential applications of computational technologies, artificial neural networks and support vector machines. Jordanski et al [36] used machine learning and MLP neural network and MLR, as well as a GCRF model to predict relationships between geometric and hemodynamic parameters for predicting WSS patterns for real models of the carotid artery and an aortic aneurysm. By comparing two models with FEM simulations, they found that GCRF can improve accuracy on both AAA and carotid bifurcation models. Gaoyang et al [37] investigated 3D cardiovascular hemodynamics of coronary artery before and after bypass using deep learning. Thy trained a deep neural network using a patient specific synthetic data from CFD simulation. Bikia et al. [38] estimated aortic systolic pressure from cuff-pressure and pulse wave velocity using regression analysis. They used Random Forest, Support Vector Machine, Ridge and Gradient Boosting for this aim. Yang et al. [39] introduced a deep learning model to parameter estimation in cardiovascular hemodynamics using convolutional neural network and fully connected neural network. They developed a model to map the nonlinear relationship between measurements including pressure waveforms, heart rate and pulse transit time, and the parameters to be estimated. Yevtushenko et al. [40] studied the proof of concept of an image-based neural network to determine the hemodynamics for patients with aortic coarctation in a centerline aggregated form. Fossan et al. [41] used a neural network to predict pressure losses in coronary arteries.

In this study, we have developed a machine that can accurately predict hemodynamic characteristics such as pressure drop, velocity, and FFR of the LAD artery from left coronary systems. It has been shown that deep neural networks can be a promising alternative to computational fluid dynamics for calculating pressures and velocities in coronary arteries and, as future works, to cardiovascular problems. Geometrical features of artery and velocity boundary conditions were used as input features of the neural network, and the pressure and velocity distributions over time, obtained from FEM-CFD simulation were used as the outputs of the network. The model was tested using three patient-specific geometries, and the results showed that our proposed model can accurately predict pressure and velocity distribution in the artery domain.

2. Methods

2.1. Governing Equations, Hemodynamics and boundary conditions

The local-parameterized models of basic shapes of the arteries, such as a bifurcation with stenosis (LAD) and three realistic reconstructed straight arteries extracted from 4-D MRI representative of the left coronary artery have been used in this study. The flow rate varied within the range of 0.8 to 4cc/s. Blood can be considered as Newtonian fluid in arteries whose internal diameter is large compared with the size of the red blood cells. Furthermore, for the arteries with a large internal diameter (larger than 1 mm) blood can be considered as a homogeneous fluid with a viscosity independent of the velocity gradient [42, 43]. The computational fluid dynamics based on the finite element method (FEM) has been used to analyze blood flow in the arteries to solve the Navier Stokes equations. The flow has been considered incompressible, Newtonian, pulsatile, and transient and for the simulation at least 3 heart cardiac cycles have been solved, (The 3-rd cardiac cycle has been considered for analysis). The Navier-Stocks equations for the simulation of the blood flow in the cardiovascular system have been given as follow:

$$\frac{\partial u}{\partial t} + u\nabla u = -\frac{1}{\rho}\nabla p + \mu \nabla^2 u \tag{1}$$

$$\frac{\partial u}{\partial t} + u\nabla(u - u_{wall}) = -\frac{1}{\rho}\nabla p + \mu \nabla^2 u \tag{2}$$

Where u represents the relative fluid velocity vector, P is pressure and µ is the viscosity. By considering blood as a Newtonian fluid, a linear correlation exists between viscous stress and the corresponding strain rate. The main boundary conditions of the computational modeling have been considered as below:

- Inlet boundary condition

Mass flow boundary conditions have been used to provide a prescribed flow rate and mass flux distribution at an inlet. For the pulsatile flow, a real profile for the inlet has been used. The frequency of heart rate has been considered 60 bpm and amplitude has been increased linearly from 0.8-4(cc/s) according to the realistic pulse model.

- Outlet boundary condition

At the outlets, a pulsatile pressure boundary condition has been used.

- Wall boundary condition

No-slip boundary condition has been considered on the wall and the fluid velocity at the wall has been set to zero.

- Fluid properties

Incompressible, Newtonian, no mesh deformation with dynamic viscosity $\nu = 0.003528$ Pa.s and density $\rho = 1060$ kgm$^{-3}$.

For the simulation, first, hexahedral grids have been generated on the geometry and a stabilized finite element model has been used. To be sure about the results accuracy and independency from pulse time in transient flow and grid sizes, a grid and time independency test have been done before performing the simulations.

### 2.2. Stenosis and pressure drop

Stenosis as the most common type of coronary artery disease (CAD) is caused by plaques generation on the endothelial walls of the coronary that lead to a decrease in the cross-sectional area of blood supply to the myocardium. The significance of the patient-specific coronary stenosis hemodynamics and the progression threat can be evaluated by comparing the hemodynamic effects induced by flow disorders [44-46]. An important parameter in coronary arterial hemodynamics is pressure drop, because of its direct relation to the power required by the heart to supply the organs. Stenosis in the arteries causes a significant pressure drop through the artery [46-48]. A higher pressure drop will result in a considerable effort that is required by heart muscle for supplying a similar blood flow rate and in some cases, it causes diseases such as heart attack and sometimes death.

Ferydoonmehr et al [48] presented a model in which pressure drop in the artery consisted of three main effects: viscous friction, blockage part, and pulsatile part. The equation of the pressure drop has been given as below:

$$\Delta P(t) = \underbrace{\frac{128}{\pi}\frac{\mu q(t)}{D_h^4}L}_{viscous\ friction\ part} + \underbrace{\frac{k_t}{2}((\frac{A_0}{A_1})^2 - 1)\rho V|V|}_{blockage\ part} + \underbrace{k_u \rho L \frac{dV}{dt}}_{pulsatile\ part} \qquad (3)$$

Where L is the length of the artery and $k_t$ and $k_u$ are the constants determined from the experimental tests. In this study, the features have been selected in such a way that the effect of the pressure drop as a consequence of stenosis be an influencer and evident in the machine learning predictions.

### 2.3. Deep Neural Network

Artificial Neural Network (ANN) is a group of multiple perceptrons/neurons in a number of layers, inspired by biological neural networks and their structures, which can be used for the prediction of outcomes by utilizing experimental knowledge or data. ANN is also known as a Feed-Forward Neural network because inputs are processed only in the forward direction. Every ANN consists of 3 layers: Input, Hidden, and Output. The input layer accepts the inputs, the hidden layer processes the inputs, and the output layer produces the results. Essentially, each layer tries to learn certain weights. A training algorithm is needed for calibrating the network weights and other parameters as a function of deviations of the outputs provided by the network and the actual values.

In this section, we propose an ANN developed for predicting pressure and velocity in the coronary arteries. Many features determine the hemodynamic characteristics of the blood flow in an artery, such as geometrical features of the artery, blood properties, boundary conditions, etc. These features must be defined before starting a CFD analysis. However, computational costs have limited using CFD simulations in many cases. Our proposed model uses some of these features to predict the pressure and velocity profile of a point in a simple artery in a network. The first step is to choose the features that have the most effects on the pressure and velocity in a coronary artery.

Table 1. List of the features used in the training set

| Feature | Description |
|---|---|
| L | The length of the artery |
| Q | The blood flow |
| $A_{in}$ | The inlet cross-section of the artery |
| $A_{out}$ | The outlet cross-section of the artery |
| x | The distance of the desired point from the inlet |
| $A_x$ | The area of the artery at the desired point |
| $r/R_x$ | The ratio of the distance of the desired point from the centerline to the radius of the artery at the point |
| t | Time |
| $x_s$ | The distance of the stenosis from the inlet |
| $L_s$ | The length of the stenosis |
| S | The degree of stenosis |

These features were selected according to the Navier-stokes equations and Table.1 to describe the desired point in the artery domain (Fig. 1), i.e., the point at which the pressure and velocity must be determined.

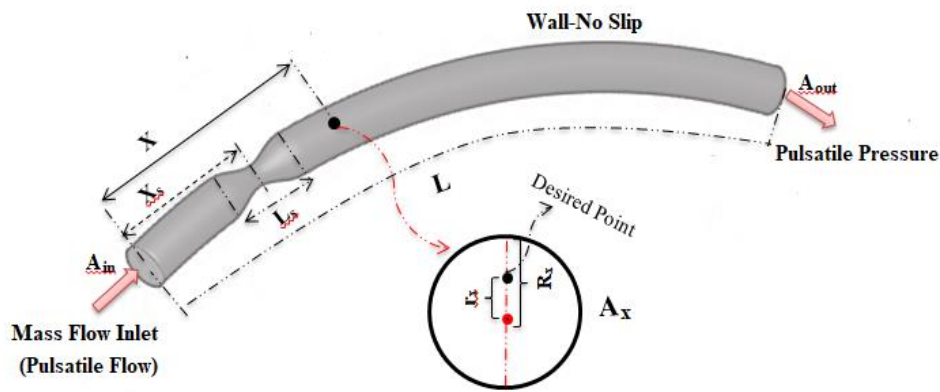

Fig.1 The geometry of the coronary artery under the study and the applied boundary conditions.

The reason we used stenosis parameters as input variables is that the stenosis can impact the flow and causes pressure drop due to energy loss [48], which means the flow after the stenosis cannot be described just by other features. The pressure boundary condition was not used in the features because obtaining precise clinical data for pressure is more difficult than obtaining velocity information. 4D MRI is a reliable and easy way to acquire blood velocity information; however, determining some important clinical factors such as FFR which is related to the pressure data needs some clinical invasive tools. So, one of the advantages of our proposed method is that it can be used to obtain pressure information in coronary arteries only by using geometry and flow features. Figure 2 illustrates the schematic of the Neural Network with input and output variables.

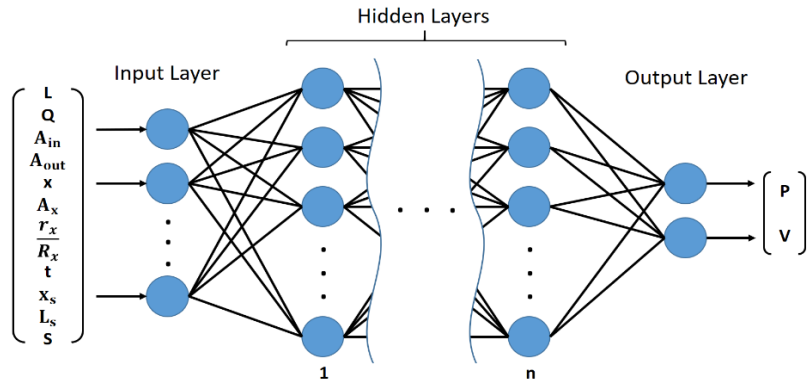

Fig. 2. The schematic Neural network with input and output variables

The model can be used to predict pressure and velocity in coronary arterial networks. It is just needed to specify the features for each artery in the arterial network and then our model can predict the output variables. In the next section, we demonstrate that how the model works. We used synthetic data which was obtained from CFD analysis to train our model and then examine the performance of the model in three real coronary artery geometries.

## 3. Results

In this section, the trained model, its parameters, and the results will be discussed. The synthetic data was generated using CFD ABAQUS simulations. Over 120 geometries with different features including the length, input, output cross-sections, and with one or multiple stenosis with different severities were modeled. The data was sampled from the ranges shown in table 2. The pressure and velocity profiles of the arteries' nodes from the CFD simulations are the output variables and the parameters shown in table 2 are the input variables or the features.

Table 2. List of the features as well as their levels used in the training set

| Feature | L (mm) | Q (cc/s) | $A_{in}$ (mm$^2$) | $A_{out}$ (mm$^2$) | x (mm) | $A_x$ (mm$^2$) | $r_x/R_x$ | t (sec) | $x_s$ (mm) | $L_s$ (mm) | S (%) |
|---|---|---|---|---|---|---|---|---|---|---|---|
| Range | 15-60 | lumped heart model (0.8-4) | 6-15 | 6-15 | 0-L | 0.6-15 | 0-1 | 0-1 | 0-L | 4-8 | 10-90 |

We used the flow and pressure waveform of coronary arteries at rest as the boundary conditions for all of the simulations [47]. The total number of nodes (all 120 geometries) is 2,400,000. It means our dataset contained 2,400,000 samples, which were used to train the model. The model was implemented in Tensor flow, version 2.4.1. Seventy percent of the data was used in the training set and 30% was used for the testing set. The model parameters and their values can be seen in table 3.

Table 3. The model parameters and their values

| Parameter | Number of Hidden Layers | Number of Neurons per Hidden Layer | Learning Rate | Optimizer | Batch Size | Number of Epochs | Nonlinearity (Activation Function) |
|---|---|---|---|---|---|---|---|
| **Value** | 2 | 64 | 0.001 | RMSprop | 10 | 100 | Relu |

The mean squared error (MSE) is used as loss function:

$$Loss(y, \hat{y}) = \frac{1}{N} \sum_{i=1}^{N} (y - \hat{y}_i)^2 \qquad (4)$$

The model's training progress for the pressure and velocity is shown in Fig. 3. It is based on mean absolute error (MAE) against the epochs. The model's performance was evaluated after the training process was overusing the test sets. Figure 4 depicts the error distribution of the test sets prediction, and Fig. 5 shows the pressure and velocity values in the testing set predicted by the model versus their reference data. One can see the model captured the pressure values more precisely. It means the pressure behavior over the features is less complicated and so more predictable.

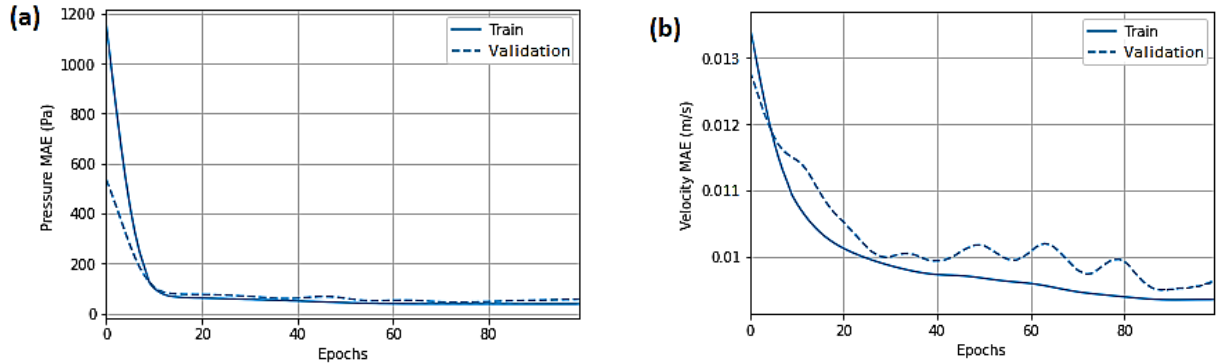

Fig. 3. Mean absolute error (MAE) during the training process for (a) Pressure and (b) Velocity.

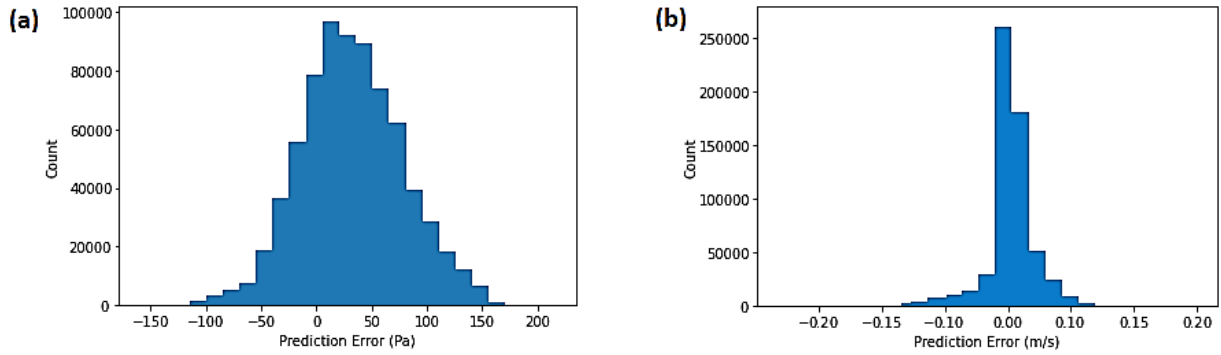

Fig. 4. Error distribution in predicting the testing set data by the model for (a) Pressure and (b) Velocity.

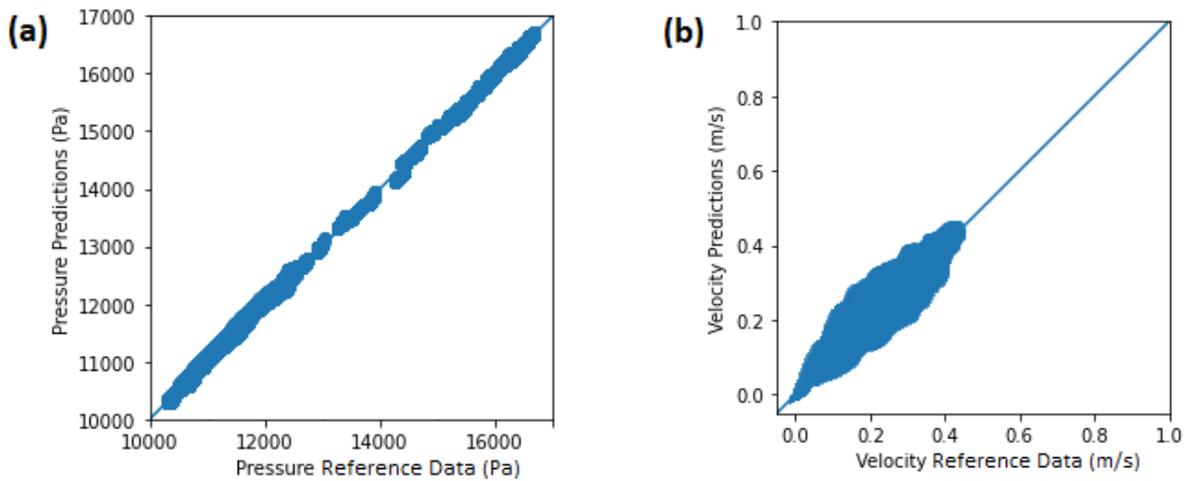

Fig. 5. Predicted values versus reference data for (a) Pressure and (b) Velocity.

The proposed method was further tested for three different cases. Cases 1 and 2 are two real geometries of LAD artery belong to two different patients and the third case is a LAD bifurcation, which is created using the point cloud method. As shown in Fig. 6, the two simple LAD arteries have stenosis that are less than 50% and the bifurcation one has one 65% stenosis in one of its branches.

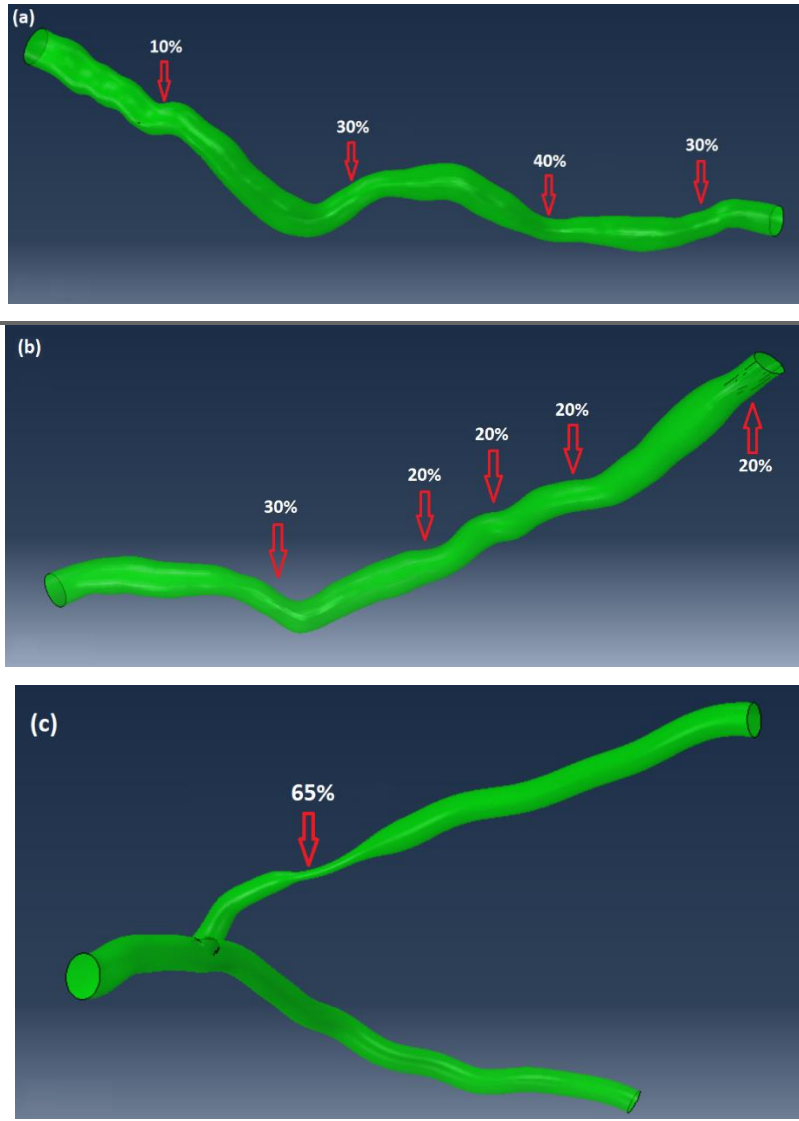

Fig. 6. The geometries used to validate the proposed method. (a) Case 1: LAD artery with a 20%, two 30%, one 40%, and one 30% stenosis. (b) Case 2: LAD artery with one 30% and four 20% stenosis. (c) Case 3: LAD bifurcation with one 65% stenosis.

For the first two cases, we used pulsatile boundary conditions for the LAD artery, i.e., flow velocity for the inlets and pressure waveform for the outlets at rest. In the third case (bifurcation), the lumped parameter heart model has been coupled for the inlet, and lumped parameter coronary vascular model has been used at the outlets [47]. The linear hexahedral mesh was used for all simulations and a mesh independency analysis was carried out to ensure the accuracy of the results. The time step for each simulation was 1 second. Figure 7 shows the mesh structure and distribution created for the bifurcation case.

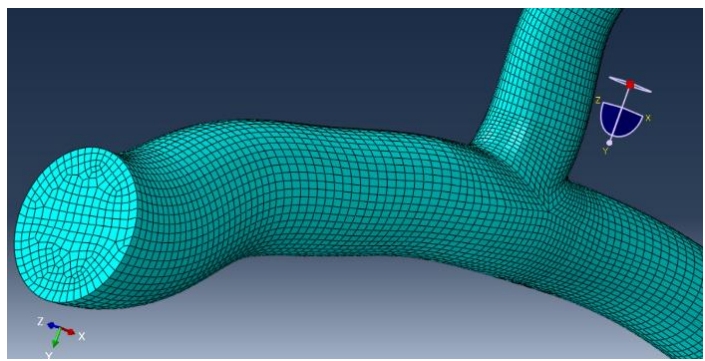

Fig. 8. An example of mesh structure and distribution
generated for the bifurcation case.

Velocity vectors and pressure distributions of the three cases at time = 0.5 s are depicted in Figs. 8 – 10. According to the figures, the pressure has been decreased from the inlet towards the outlet. The velocity contour and vectors in the presented arteries show a maximum gradient in the stenosis area (30% stenosis in case 2 and 65% stenosis in bifurcation).

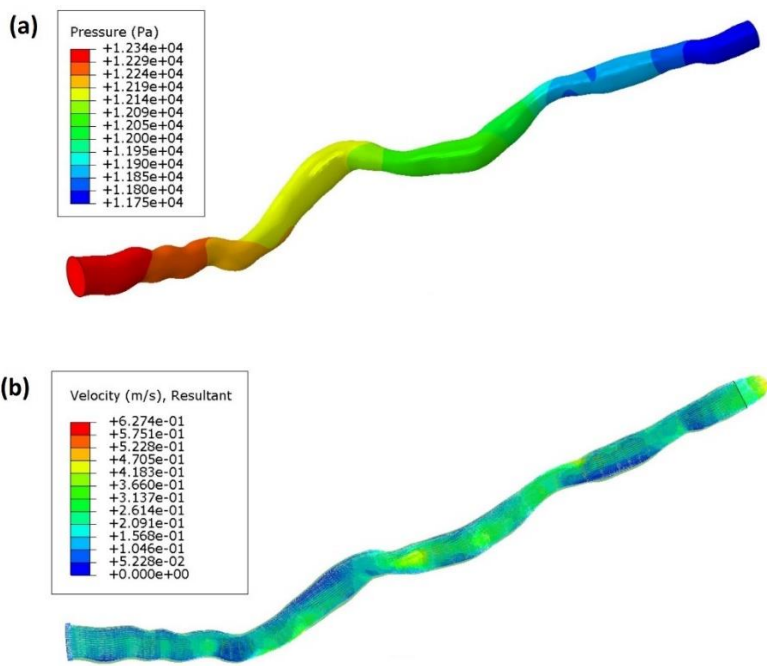

Fig. 8. (a) Pressure and (b) velocity distribution of the case 1 at t = 0.5

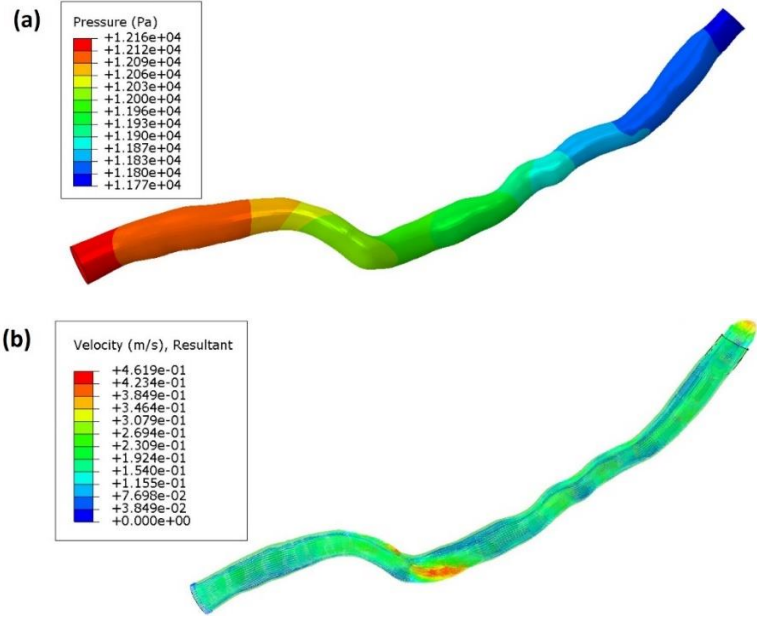

Fig. 9. (a) Pressure and (b) velocity distribution of the case 2 at t = 0.5

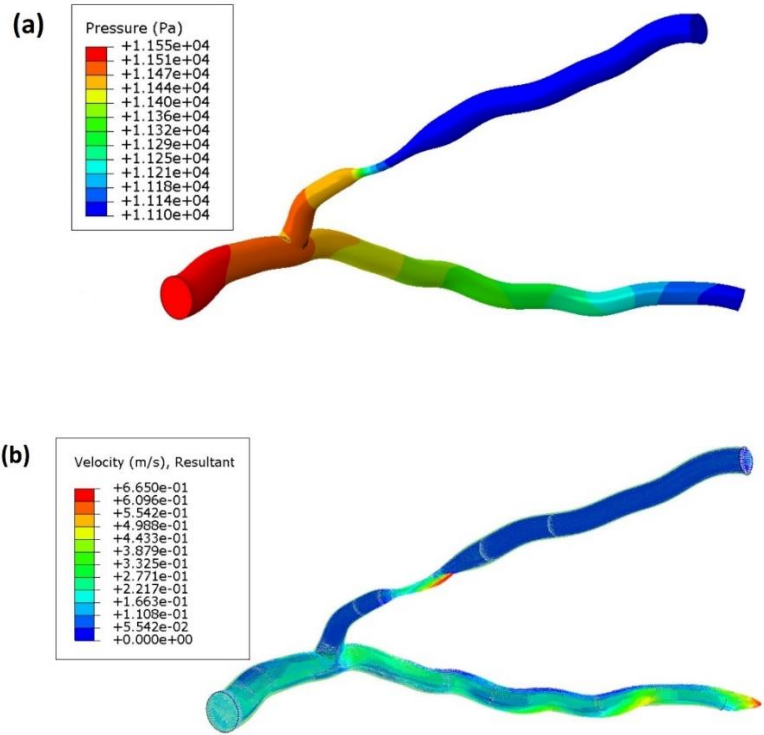

Fig. 10. (a) Pressure and (b) velocity distribution of the case 3 at t = 0.5

After extracting the data for pressure and velocity of three cases for different nodes in the domain, we used the pre-trained deep neural network to predict the pressures and velocities of the elements. The model can predict pressure and velocity in any node in the fluid domain. We picked three nodes in cases 1 and 2, and six nodes in case 3. It is worth mentioning that case three itself consists of three arteries. Figure 11 illustrates the nodes we used to examine the performance of our methods for the three cases.

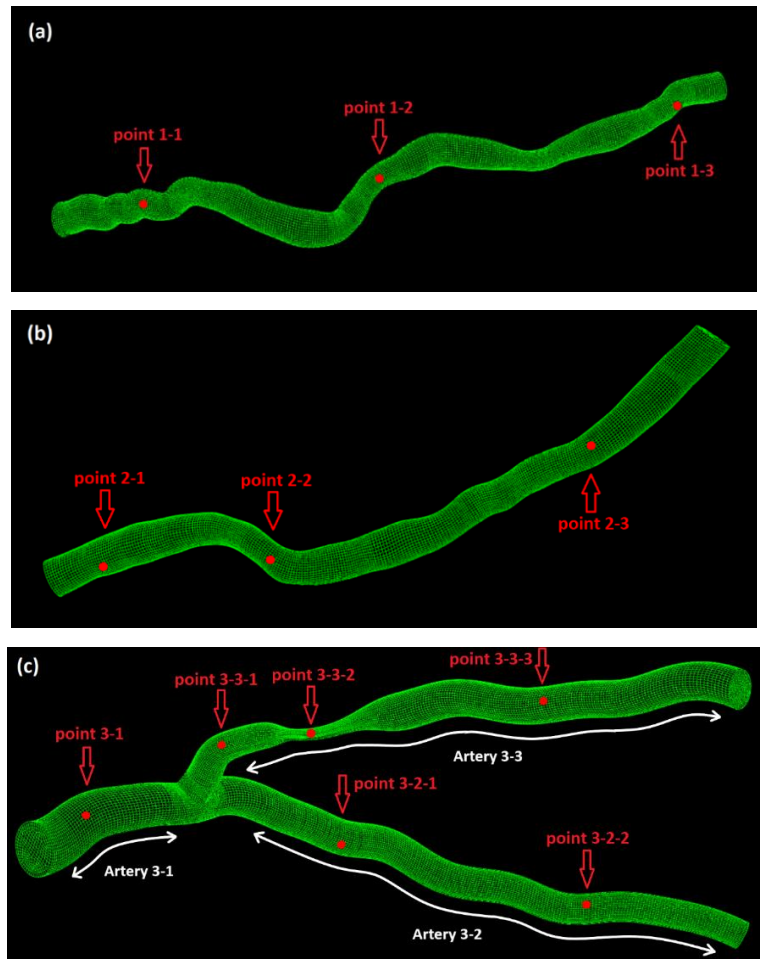

Fig. 11. The nodes used to examine the proposed method for three cases: (a) case 1, (b) case 2, (c) case 3. The predicted pressure and velocity profiles (ML) compared to the reference data (CFD) for the points specified in Fig. 11 are depicted in Fig. 12-14.

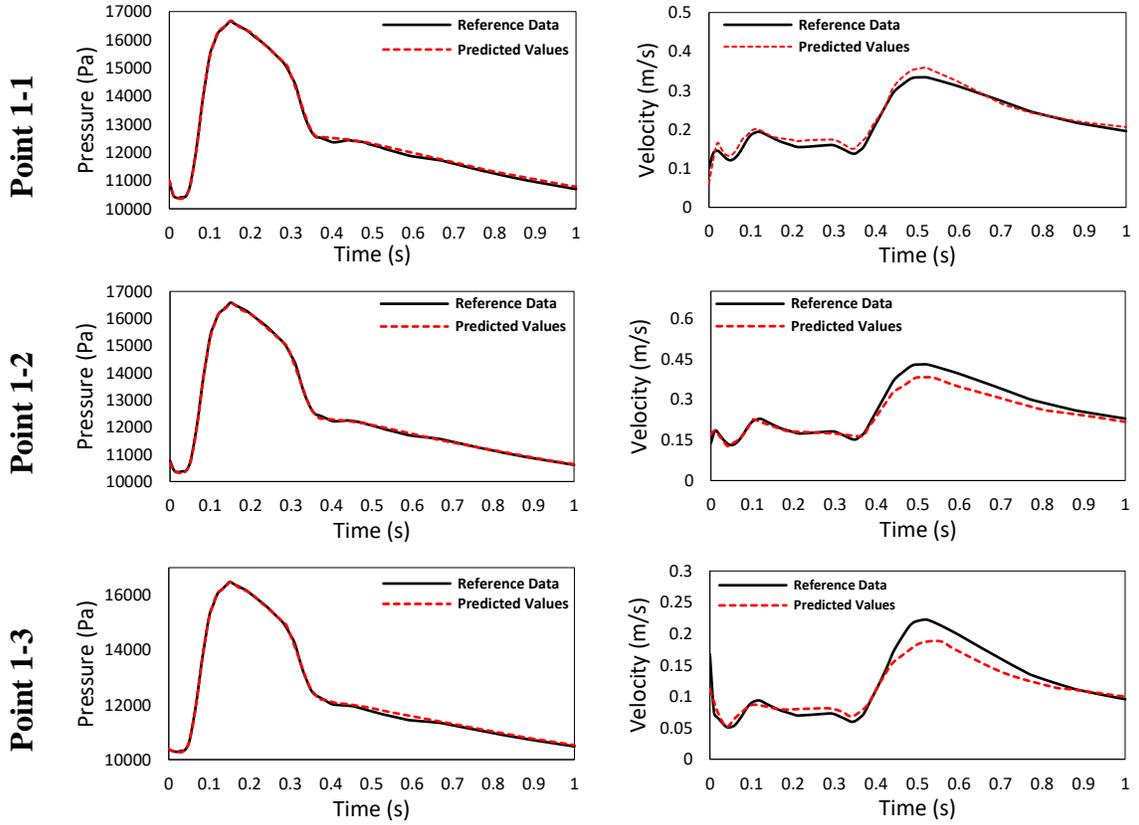

Fig. 12. Pressure and velocity profiles for three points in case 1 for one pulse (1 sec).

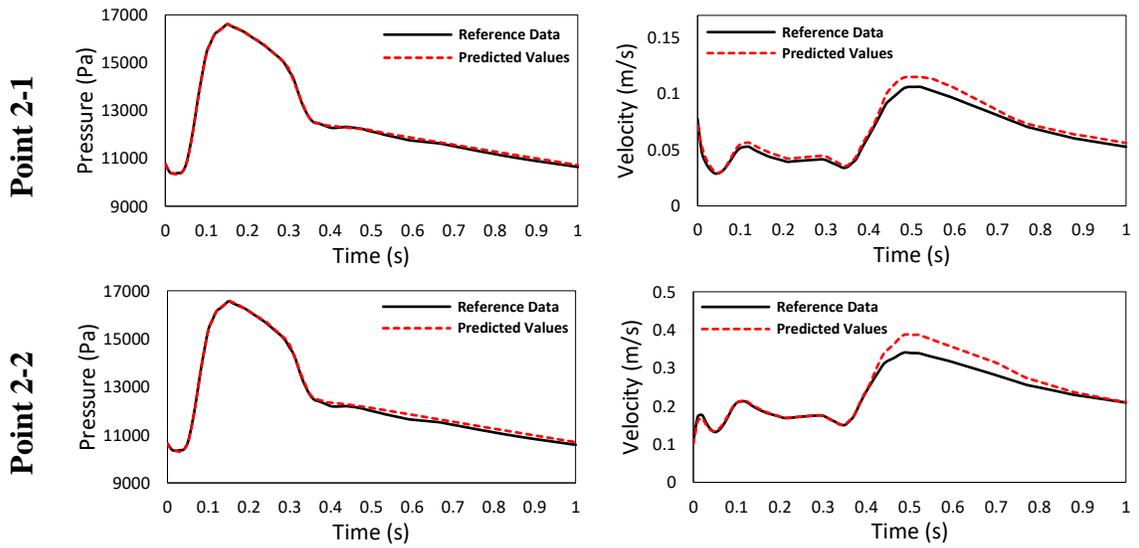

**Point 2-3**

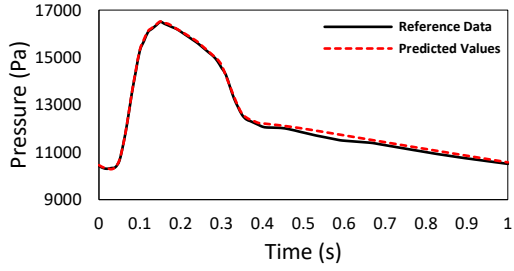
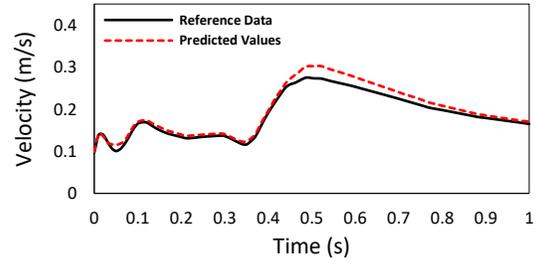

Fig. 13. Pressure and velocity profiles for three points in case 2 for one pulse (1 sec).

**Point 3-1**

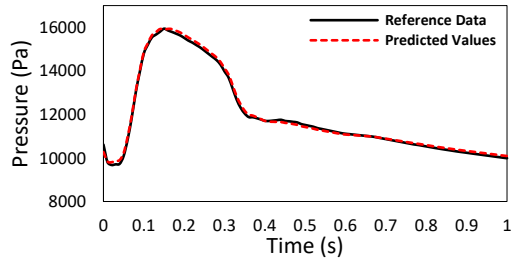
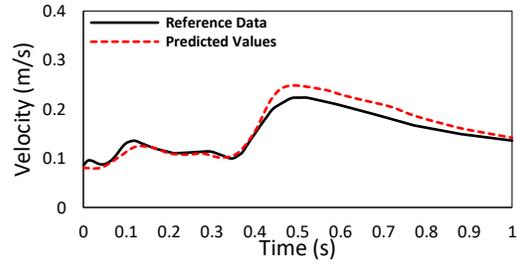

**Point 3-2-1**

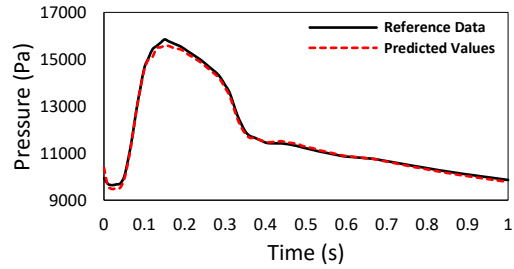
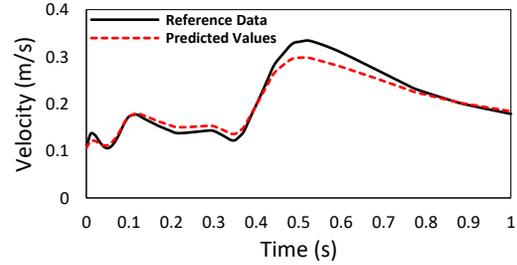

**Point 3-2-2**

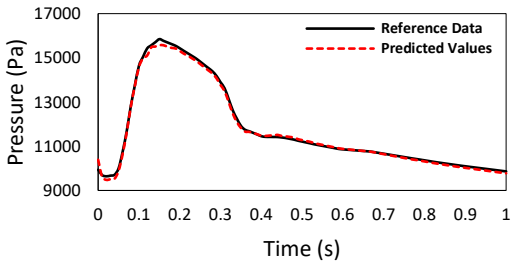
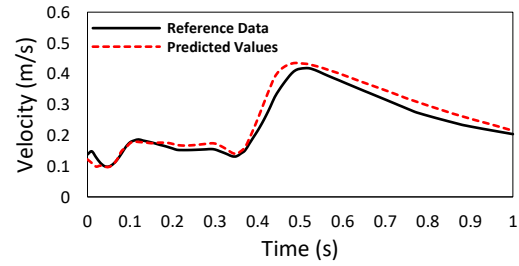

**Point 3-3-1**

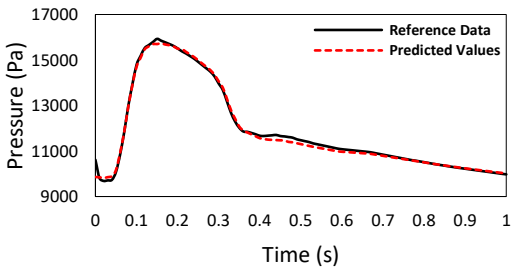
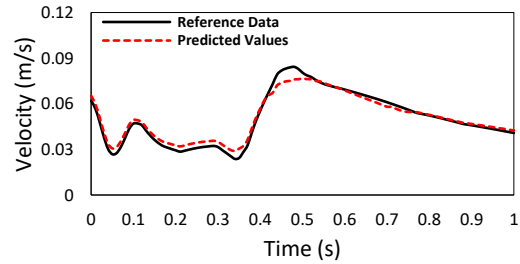

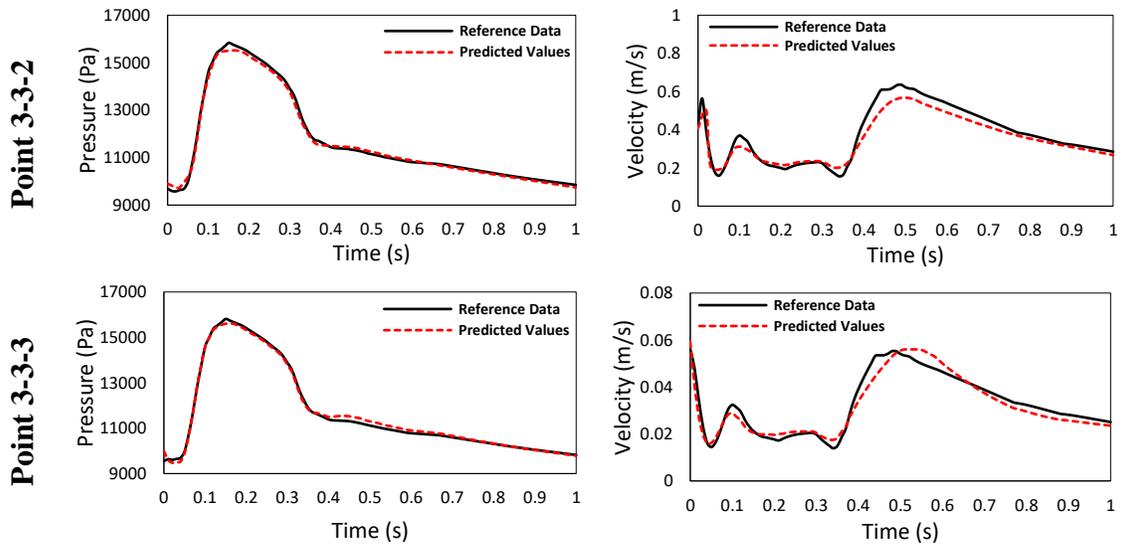

Fig. 14. Pressure and velocity profiles for six points in case 3 for one pulse (1 sec).

Fig 15 shows the mean velocity profile prediction for three cross-sections at points 1-2, 2-2 and 3-1. As one can see, the accuracy of mean velocity prediction is higher than the velocity prediction in a point.

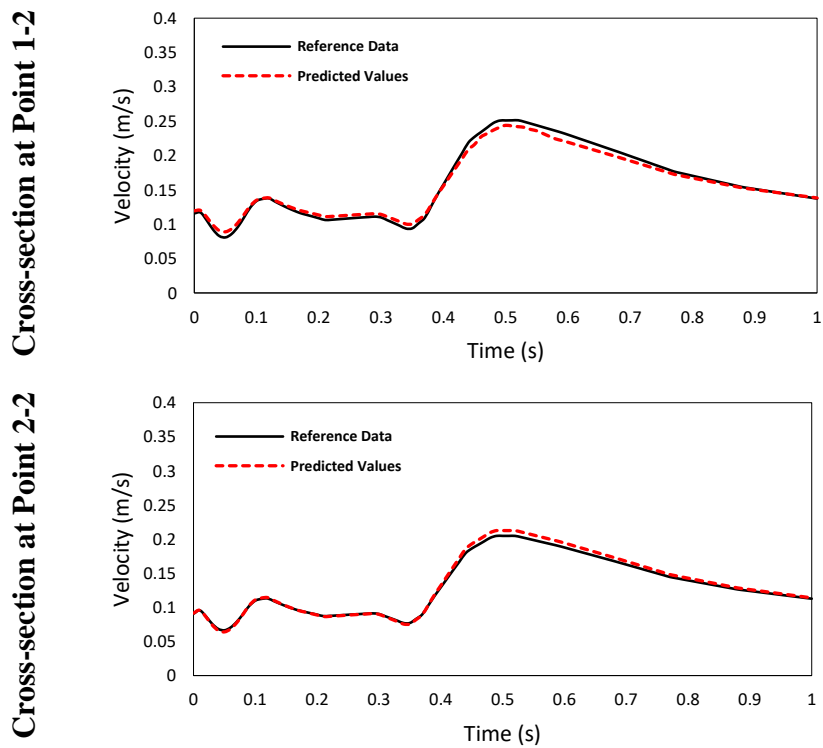

**Cross-section at Point 3-1**

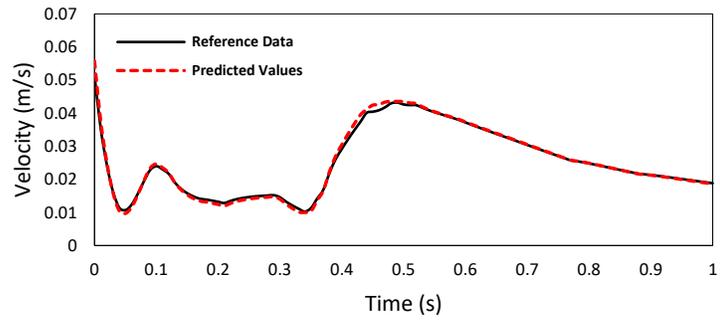

Fig. 15. Mean velocity profile for three cross-sections in three cases for one pulse (1 sec)

**Cross-section at Point 1-2**

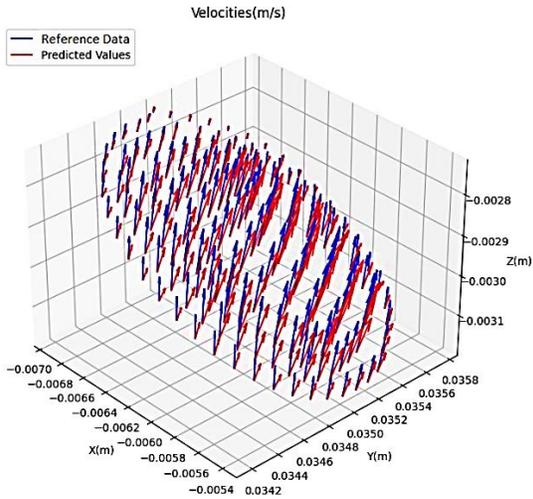 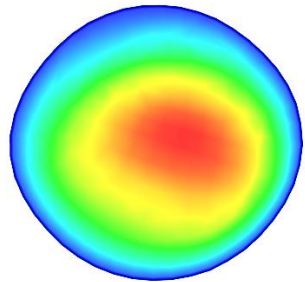 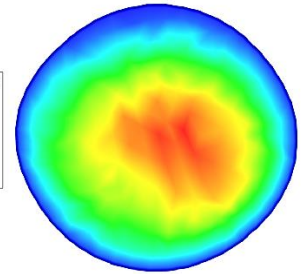

**Cross-section at Point 2-2**

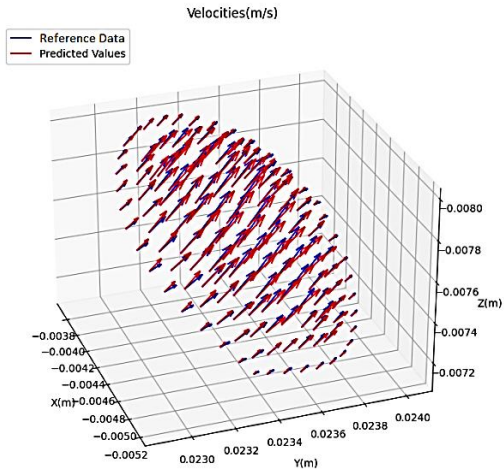 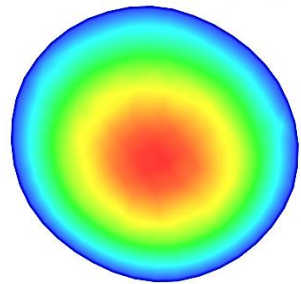 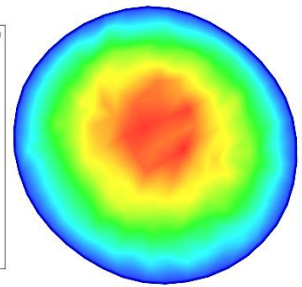

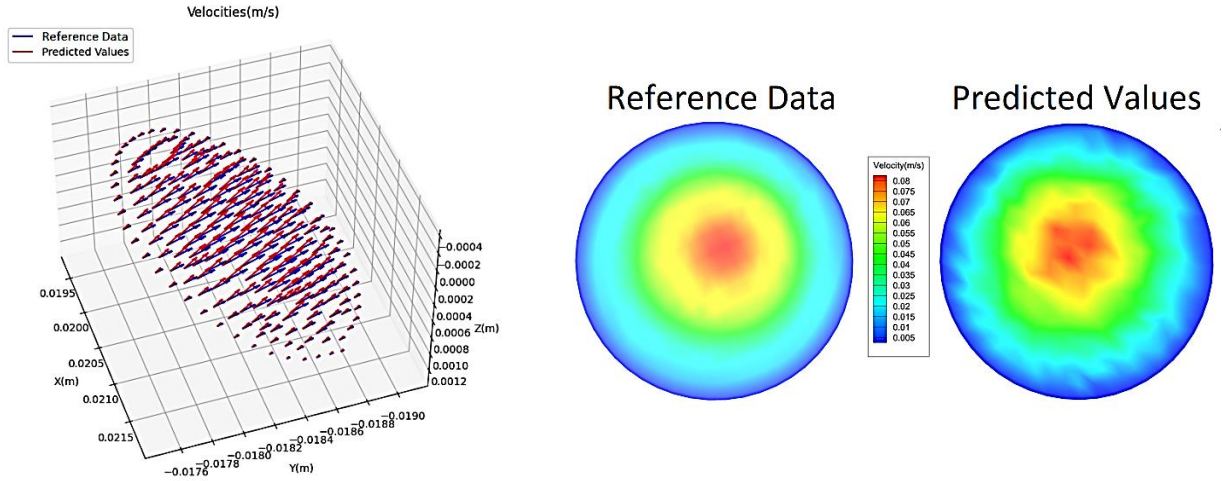

**Cross-section at Point 3-1**

Fig. 16. Velocity profiles for three cross-sections of the arteries 1, 2, and 3 at t=0.5 s.

Figs. 17 and 18 show the wall pressure along three arteries at time 0.5s and pressure drop due to stenosis at artery 3-3 respectively. The wall pressure contour through the stenosis for the artery 3-3 is depicted in the Fig. 19.

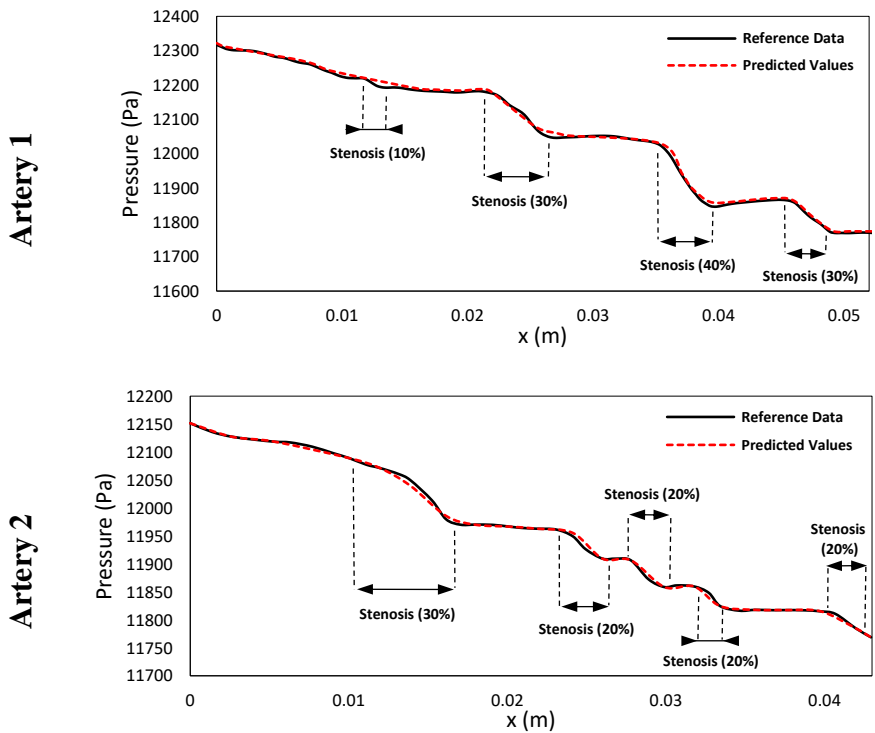

**Artery 1**

**Artery 2**

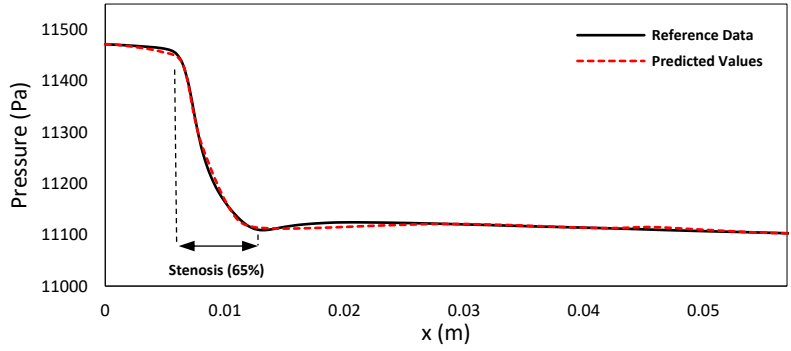

Fig. 17. Wall pressure along the arteries 1, 2, and 3-3 at t=0.5 s

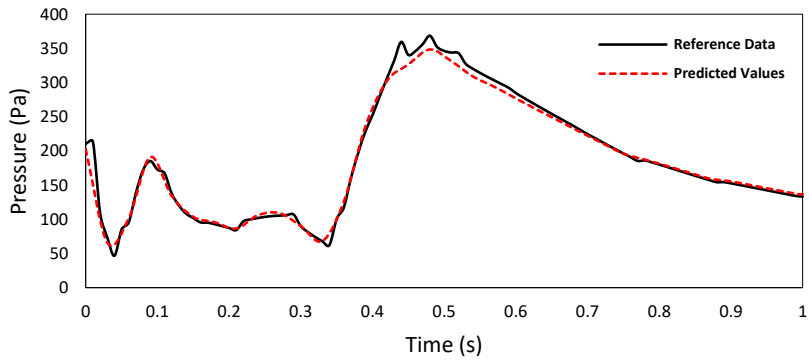

Fig. 18. The pressure drop due to stenosis in the artery 3-3 for one pulse (1 sec)

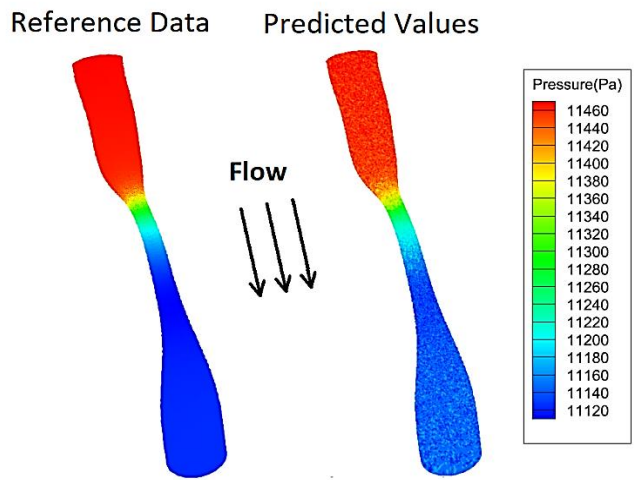

Fig. 19. The pressure through stenosis in the artery 3-3 at t=0.5 s

## 4. Discussion

As shown in Figs. 12-14, the pressure profiles predicted by the trained model are in good agreement with the reference data for all cases. However, the accuracy of the model is slightly lower at the velocity peaks where the points with high velocities may not be well captured by the model. That is, the more the velocity increases, the more unpredictable the behavior becomes. On the other hand, our model is three-dimensional and our goal was to predict every point in the domain where the variations in velocities over the features (especially r/R and $A_x$) are more drastic than pressure. In a cross-section, the velocity varies from 0 to its maximum value so the model needs much more data points for more precise predictions. Furthermore, the differences between the real geometries and the geometries used to train the model, especially where there are extreme distortions and instantaneous curvatures, may lead to this slight deviation of the predicted velocity values from their real ones at those cross-sections.

Despite the velocity at every single point in the domain, the mean velocity in each cross-section can be predicted more accurately by the model. The reason is that the mass conversation equations in the domain are satisfied in every cross-section. Considering the blood as an incompressible fluid, the flow can be constant at every cross-section in an artery at a certain moment. Therefore, at some cross-sections in real geometries, in which the artery is not straight, the velocity distribution may be different from the distribution at a cross-section with the same features in a straight artery, however, the mean velocity at two cross-sections are the same. Figure 15 shows that our model, which was trained by different straight arteries, can precisely predict the mean velocity at each cross-section in the domain of the real geometries.

The velocity profiles in three cross-sections of the arteries are depicted in Figure 16. The velocity magnitude in each node is in good agreement with the reference data (simulation). However, there is a slight difference between the velocity vector directions of the predicted values and reference data. The reason is that in the ANN model the velocity vectors are considered to be along with the direction of the artery centerline.

The magnitude of the pressure drop due to stenosis can be used as an index to determine the severity of stenosis. Our model can predict the pressure of every point in the domain based on geometrical features and inlet velocity, thus the pressure drop can be predicted. In Fig. 17, the wall pressure along the arteries 1, 2, and 3-3 at t=0.5 s which were predicted by our proposed model is compared with those of real values. Figure 18 depicts the pressure drop in the artery 3-3, i.e., the difference between the downstream and upstream pressures of the stenosis over time. The pressure drop can be precisely predicted by the proposed model.

One of the most important indices to determine the severity of stenosis is fractional flow reserve (FFR). FFR represents the ratio of flow in the stenosed artery at hyperemia with the highest flow rate to the maximum reachable blood flow rate in the same branch in a healthy condition. The ratio of the mean distal coronary artery pressure to the aortic pressure during maximal vasodilatation also represents FFR [29, 44]. FFR can be evaluated based on the pressure that the model predicts for every point in the artery. According to the pressure drop predicted in the artery 3-3 due to the 65% stenosis, the FFR is 0.9, which is in good agreement with the FFR calculated based on the simulation (0.91). The predicted and real wall pressure through the stenosis is shown in Figure 19 for case 3.

The average precision of the pressure and velocity through the heart cycle is 98.7 and 93.2 respectively. The precision was calculated using the predicted values and reference data during the heart cycle for the test points in the all three cases.

## 5. Conclusion

In this study, we proposed a machine learning approach to predict the hemodynamic characteristics of coronary arteries. It was shown that deep neural networks can be a promising alternative to computational fluid dynamics for calculating pressures and velocities in coronary arteries and, as future works, to cardiovascular problems. Geometrical features of artery and velocity boundary conditions were used as input features of the neural network, and the pressure and velocity distributions over time, obtained from CFD simulations, were used as the outputs of the network. The model was tested on three patient-specific geometries, and the results showed that our model can predict pressure and velocity distributions in the artery domains with high accuracy. The prediction was more precise for the pressure compared to that of the velocity. Nonetheless, the prediction of the velocity at each point would be more accurate if more data points and features, for example, artery's curvature radius were available for the training. This brings the artery distortions into account as an affecting factor on the velocity distribution. Furthermore, although, as a cost and time-efficient method, the model was trained for coronary arteries, however, it can be easily retrained for other types of arteries in the body using a new dataset. One of the most important results of the model is to predict the pressure drop due to the existence of stenosis in the artery. For example, Atherosclerosis is serious damage that should be addressed using new approaches such as machine learning. Since the proposed model can predict the pressure at all points in the domain, one can calculate FFR for every point in an artery as an important factor to quantify the severity of stenosis. Last but not least, the proposed method is independent of the source of data, i.e., it can be trained using any type of dataset. Therefore, the accuracy of the predictions can be further improved once real and more comprehensive clinical data are used to train the model.

**References:**

[1] T. Heldt, R. Mukkamala, G.B. Moody, R.G. Mark, CVSim. An open-source cardiovascular simulator for teaching and research, Open Pacing Electrophysiol. Ther. J., 3, pp. 45-54, 2010.

[2] A. Goodwin, W.L. van Meurs, C.D. Sá Couto, J.E. Beneken, S.A. Graves, A model for educational simulation of infant cardiovascular physiology, Anesth. Analg., 99 (6), pp. 1655-1664, 2004.

[3] G. Ferrari, M. Kozarski, K. Zieliński, L. Fresiello, A. Di Molfetta, K. Górczyńska, K.J. Pałko, M. Darowski, A modular computational circulatory model applicable to VAD testing and training, J. Artif. Organs, 15: 32-43, 2012.

[4] M. Tokaji, S. Ninomiya, T. Kurosaki, K. Orihashi, T. Sueda, An educational training simulator for advanced perfusion techniques using a high-fidelity virtual patient model Artif. Organs, 36 (12). 1026-1035, 2012.

[5] A. Stefanovska, Physics of the human cardiovascular system, Contemp. Phys. 40 (1), 31–55, 1999.

[6] Y. Ma, J. Choi, A. Hourlier-Fargette, Y. Xue, H.U. Chung, J.Y. Lee, X. Wang, Z. Xie, D. Kang, H. Wang, et al., Relation between blood pressure and pulse wave velocity for human arteries, Proc. Natl. Acad. Sci. 115 (44),11144–11149, 2018.

[7] P. Reymond, F. Merenda, F. Perren, D. Rufenacht, N. Stergiopulos, Validation of a one-dimensional model of the systemic arterial tree, Amer. J. Physiol. Heart Circ. Physiol. 297 (1),H208–H222, 2009.

[8] M. Haris, A. Singh, K. Cai, F. Kogan, J. McGarvey, C. DeBrosse, G.A. Zsido, W.R. Witschey, K. Koomalsingh, J.J. Pilla, et al., A technique for in vivo mapping of myocardial creatine kinase metabolism, Nature Med. 20 (2),209, 2014.

[9] C.A. Figueroa, I.E. Vignon-Clementel, K.E. Jansen, T.J. Hughes, C.A. Taylor, A coupled momentum method for modeling blood flow in three-dimensional deformable arteries, Comput. Methods Appl. Mech. Engrg. 19:5685–5706, 2006.

[10] V.O. Kheyfets, L. Rios, T. Smith, et al. Patient-specific computational modeling of blood flow in the pulmonary arterial circulation. Comput Methods Programs Biomed 2015;120:88–10.

[11] C. Bertoglio, D. Barber, N. Gaddum, et al. Identification of artery wall stiffness: in vitro validation and in vivo results of a data assimilation procedure applied to a 3D fluid-structure interaction model. J Biomech, 47:1027–34, 2014.

[12] N.H. Pijls, W.F. Fearon, P.A. Tonino, et al. Fractional flow reserve versus angiography for guiding percutaneous coronary intervention in patients with multivessel coronary artery disease: 2-year follow-up of the FAME (Fractional Flow Reserve Versus Angiography for Multivessel Evaluation) study. J Am Coll Cardiol,56:177–84, 2010.

[13] S.J. Sonntag, W.Li, M. Becker, et al. Combined computational and experimental approach to improve the assessment of mitral regurgitation by echocardiography. Ann Biomed Eng ,42:971–85, 2014.

[14] B. Balakrishnan, A. R. Tzafriri, P. Seifert, A. Groothuis, C. Rogers, E. R. Edelman, Strut position, blood flow, and drug deposition: implications for single and overlapping drug-eluting stents, Circulation 111 (22),2958–2965, 2005.

[15] M. Gay, L. Zhang, W. K. Liu, Stent modeling using immersed finite element method, Comput Methods Appl Mech Eng 195, 4358–4370, 2006.

[16] W. Wu, M. Qi, X. Liu, D. Yang, W. Wang, Delivery and release of nitinol stent in carotid artery and their interactions: a finite element analysis, J Biomech, doi:10.1016/j.jbiomech.2007.02.024.

[17] W.X.Chen, Eric K.W.Poon, V. Thondapu, N. Hutchins, P. Barlisb, A. Ooi, Haemodynamic effects of incomplete stent apposition in curved coronary arteries, Journal of Biomechanics, 63- 3: 164-173,2017.

[18] T. Xiaopeng, A. Sun, L. Xiao, F. Pu, D. Xiaoyan, H. Kang, Y. Fan, Influence of catheter insertion on the hemodynamic environment in coronary arteries, Medical Engineering & Physics, 38- 9, 946-951, 2016.

[19] S. Numata, K. Itatani, H. Kawajiri, S. Yamazaki, K. Kanda, H. Yaku, Computational fluid dynamics simulation of the right subclavian artery cannulation, The Journal of Thoracic and Cardiovascular Surgery, 154- 2, 480-487, 2017.

[20] G. Rigatelli, M. Zuin, F. Dell'Avvocata, D. Vassilev, R. Daggubati, T. Nguyen, N.Van Viet Thang, N. Foinh, Evaluation of coronary flow conditions in complex coronary artery bifurcations stenting using computational fluid dynamics: Impact of final proximal optimization technique on different double-stent techniques, Cardiovascular Revascularization Medicine, Volume 18, Issue 4, Pages 233-240, 2017.

[21] D. Hardman, B. J. Doyle, S. IK Semple, J. MJ Richards, D. E Newby, W. J Easson and P. R Hoskins,On the prediction of monocyte deposition in abdominal aortic aneurysms using computational fluid dynamics, Proc ImechE Part H: J Engineering in Medicine ,227(10) 1114–1124, 2013.

[22] D. Hardman, B.J. Doyle, S.I. Semple, et al. On the prediction of monocyte deposition in abdominal aortic aneurysms using computational fluid dynamics. Proceedings of the Institution of Mechanical Engineers, Part H: Journal of Engineering in Medicine. 227(10):1114-1124, 2013.

[23] K.R. Foster, R. Koprowski, J.D. Skufca,Machine learning, medical diagnosis, and biomedical engineering research: Commentary. BioMedical Engineering OnLine. 13:94, 2014.

[24] K. Kourou, T.P. Exarchos, K.P. Exarchos, M.V. Karamouzis, D.I. Fotiadis. Machine learning applications in cancer prognosis and prediction. Computational and Structural Biotechnology Journal. 13:8–17, 2015.

[25] M.V. Albert, K. Kording, M. Herrmann, A. Jayaraman. Fall classification by machine learning using mobile phones. PloS One. 7:e36556, 2012.

[26] M. Motwani, D. Dey, D.S. Berman, G.Germano, S.Achenbach, M.H. Al-Mallah, D.Andreini, M.J. Budoff, F.Cademartiri, T.Q. Callister, H.J. Chang, K. Chinnaiyan, B.J.Chow, R.C. Cury, A. Delago, M.

Gomez, H. Gransar, M. Hadamitzky, J. Hausleiter, N. Hindoyan, G. Feuchtner, P.A. Kaufmann, Y.J. Kim, J. Leipsic, F.Y. Lin, E. Maffei, H. Marques, G. Pontone, G. Raff, R. Rubinshtein, L.J. Shaw, J. Stehli, T.C. Villines, A. Dunning, J.K. Min, P.J. Slomka. Machine learning for prediction of all-cause mortality in patients with suspected coronary artery disease: a 5-year multicenter prospective registry analysis. Eur Heart J. 2017;38:500–507.

[27] R. Arsanjani, Y. Xu, D. Dey, V. Vahistha, A. Shalev, Nakanishi R, Hayes S, Fish M, Berman D, Germano G, Slomka PJ. Improved accuracy of myocardial perfusion SPECT for detection of coronary artery disease by machine learning in a large population. J Nucl Cardiol. 20:553–562,2013.

[28] S. Sankarana, L. Grady, Charles A. Taylorc, Impact of geometric uncertainty on hemodynamic simulations using machine learning, Comput. Methods Appl. Mech. Engrg. 297:167–190, 2015.

[29] L. Itu, S. Rapaka, T. Passerini, B. Georgescu, C. Schwemmer, M. Schoebinger, T. Flohr, P. Sharma, and D. Comaniciu, A machine-learning approach for computation of fractional flow reserve from coronary computed tomography, J Appl Physiol 121: 42–52, 2016.

[30] T.H. Wu, G.K. Pang and E.W. Kwong, Predicting Systolic Blood Pressure Using Machine Learning,7th International Conference on Information and Automation for Sustainability, Colombo, Sri Lanka, pp. 1-6, 2014.

[31] A. Narang, V. Mor-Avi, A. Prado, V. Volpato, D. Prater, G.Tamborini, L. Fusini, M. Pepi, N.Goyal, K. Addetia, A. Gonc¸alves, A.R. Patel, and R.M. Lang, Machine learning based automated dynamic quantification of left heart chamber volumes, European Heart Journal – Cardiovascular Imaging, 1–9, 2018.

[32] C. Tesche, C. N. De Cecco, S. Baumann, M. Renker, T.W. McLaurin, T.M. Duguay, R.R. Bayer, D.H. Steinberg, K.L. Grant, C. Canstein, C. Schwemmer, M. Schoebinger, L.M. Itu, S. Rapaka, P.Sharma, U. Joseph Schoep, Coronary CT Angiography–derived Fractional Flow Reserve: Machine Learning Algorithm versus, Computational Fluid Dynamics Modeling, Radiology,288:64–72, 2018.

[33] A. Coenen, Y.H. Kim, M. Kruk, C. Tesche, J. De Geer, A. Kurata, M.L. Lubbers, J. Daemen, L. Itu, S. Rapaka, P. Sharma, C. Schwemmer, A. Persson, U.J. Schoepf, C. Kepka, D. Hyun Yang, Nieman K. Diagnostic Accuracy of a Machine-Learning Approach to Coronary Computed Tomographic Angiography-Based Fractional Flow Reserve: Result From the MACHINE Consortium. Circ Cardiovasc Imaging. 2018 Jun;11(6):e007217. Doi: 10.1161/CIRCIMAGING.117.007217. PMID: 29914866.

[34] G. Kissas, Y. Yang, E. Hwuang, W. R. Witschey, J. A. Detre, P. Perdikaris, Machine learning in cardiovascular flows modeling: Predicting arterial blood pressure from non-invasive 4D flow MRI data using physics-informed neural networks, Computer Methods in Applied Mechanics and Engineering 358,112623, 2018.

[35] M. Cilla, J. Martınez, E. Pena, and M. Angel Martınez, Machine Learning Techniques as a Helpful Tool Toward Determination of Plaque Vulnerability, IEEE transactions on biomedical engineering. 59- 4, 2012.

[36] M. Jordanski, M. Radovic, Z. Milosevic, N. Filipovic and Z. Obradovic, "Machine Learning Approach for Predicting Wall Shear Distribution for Abdominal Aortic Aneurysm and Carotid Bifurcation Models," in IEEE Journal of Biomedical and Health Informatics, vol. 22, no. 2, pp. 537-544, March, 2018, doi: 10.1109/JBHI.2016.2639818.

[37] Li, Gaoyang, Haoran Wang, Mingzi Zhang, Simon Tupin, Aike Qiao, Youjun Liu, Makoto Ohta, and Hitomi Anzai. Prediction of 3D Cardiovascular hemodynamics before and after coronary artery bypass surgery via deep learning, Communications biology. 4, no. 1, 1-12, 2021

[38] Bikia, Vasiliki, Theodore G. Papaioannou, Stamatia Pagoulatou, Georgios Rovas, Evangelos Oikonomou, Gerasimos Siasos, Dimitris Tousoulis, and Nikolaos Stergiopulos. Noninvasive estimation of aortic hemodynamics and cardiac contractility using machine learning. Scientific Reports 10, no. 1, 1-17, 2020.

[39] Zhou, Yang, Yuan He, Jianwei Wu, Chang Cui, Minglong Chen, and Beibei Sun. "A method of parameter estimation for cardiovascular hemodynamics based on deep learning and its application to personalize a reduced-order model. International Journal for Numerical Methods in Biomedical Engineering. e3533, 2021.

[40] Yevtushenko, Pavlo, Leonid Goubergrits, Lina Gundelwein, Arnaud Setio, Tobias Heimann, Heiko Ramm, Hans Lamecker, Titus Kuehne, Alexander Meyer, and Marie Schafstedde. Deep Learning Based Centerline-Aggregated Aortic Hemodynamics: An Efficient Alternative to Numerical Modelling of Hemodynamics. IEEE Journal of Biomedical and Health Informatics, 2021.

[41] Fossan, Fredrik E., Lucas O. Müller, Jacob Sturdy, Anders T. Bråten, Arve Jørgensen, Rune Wiseth, and Leif R. Hellevik. Machine learning augmented reduced-order models for FFR-prediction. Computer Methods in Applied Mechanics and Engineering 384, 113892, 2021.

[42] R. H. Haynes, Physical basis of the dependence of blood viscosity on tube radius, Am. J. Physiol. 198, 1193–1200, 1960.

[43] W. Nichols, M. O'Rourke, and C. Vlachopoulos, McDonald's Blood Flow in Arteries: Theoretical, Experimental and Clinical Principles, 6th ed, 2011.

[44] C.W. Akins, B. Travis, & A.P. Yoganathan. Energy loss for evaluating heart valve performance. The Journal of Thoracic and Cardiovascular Surgery, 136(4), 820–833, 2008. doi:10.1016/j.jtcvs.2007.12.059.

[45] M.Berger, R.L. Berdoff, P.E. Gallerstein, & E. Goldberg.  Evaluation of aortic stenosis by continuous wave Doppler ultrasound. Journal of the American College of Cardiology, 3(1), 150–156. doi:10.1016/s0735-1097(84)80442-8, 1984.

[46] H.J. Kim, I.E.Vignon-Clementel, J.S.Coogan, C.A.Figueroa, K.E. Jansen, C.A. Taylor. Patient-specific modeling of blood flow and pressure in human coronary arteries. Ann Biomed Eng. 38(10):3195-209,2010.

[47] Z. Malota, J. Glowacki, W. Sadowski, M. Kostur. Numerical analysis of the impact of flow rate, heart rate, vessel geometry, and degree of stenosis on coronary hemodynamic indices. BMC Cardiovasc Disord. 2018 Jun 28;18(1):132.

[48] N. Freidoonimehr, R. Chin, A. Zander, and M. Arjomandi, An experimental model for pressure drop evaluation in a stenosed coronary artery Physics of Fluids 32, 021901 (2020).

[49] N.H. Pijls, B. De Bruyne, K. Peels, P.H. Van Der Voort, H.J. Bonnier, J. Bartunek, J.J. Koolen, Koolen JJ. Measurement of fractional flow reserve to assess the functional severity of coronary-artery stenoses. N Engl J Med, 27;334(26):1703-8, 1996.